# Large Language Models for Persian ↔ English Idiom Translation

**Sara Rezaeimanesh**[*1,2] **Faezeh Hosseini**[*2] **Yadollah Yaghoobzadeh**[1,2]

[1]School of Electrical and Computer Engineering,
College of Engineering, University of Tehran, Tehran, Iran,
[2]Tehran Institute for Advanced Studies, Khatam University, Tehran, Iran,
**Correspondence:** srezaeimanesh@ut.ac.ir, f.hosseini401@khatam.ac.ir, yyaghoobzadeh@ut.ac.ir

## Abstract

Large language models (LLMs) have shown superior capabilities in translating figurative language compared to neural machine translation (NMT) systems. However, the impact of different prompting methods and LLM-NMT combinations on idiom translation has yet to be thoroughly investigated. This paper introduces two parallel datasets of sentences containing idiomatic expressions for Persian→English and English→Persian translations, with Persian idioms sampled from our PersianIdioms resource, a collection of 2,200 idioms and their meanings, with 700 including usage examples. Using these datasets, we evaluate various open- and closed-source LLMs, NMT models, and their combinations. Translation quality is assessed through idiom translation accuracy and fluency. We also find that automatic evaluation methods like LLM-as-a-judge, BLEU, and BERTScore are effective for comparing different aspects of model performance. Our experiments reveal that Claude-3.5-Sonnet delivers outstanding results in both translation directions. For English→Persian, combining weaker LLMs with Google Translate improves results, while Persian→English translations benefit from single prompts for simpler models and complex prompts for advanced ones.[1]

## 1 Introduction

An idiom is a phrase or expression with a figurative meaning distinct from its literal interpretation. Idioms are commonly used in everyday language to convey ideas more vividly and often originate from cultural, historical, or social contexts, making them specific to particular languages or regions. Idiomatic expressions present significant challenges for NLP, particularly in translating between culturally distinct languages such as Persian and English.

Despite their prevalence in spoken language, state-of-the-art machine translation (MT) models struggle with translating idioms, often rendering them literally as compositional expressions (Raunak et al., 2023; Dankers et al., 2022). Early MT efforts attempted to address this problem using idiom dictionaries or direct substitution (Salton et al., 2014; Nagao, 1984). However, idioms evolve and vary by context, and even identical meanings can require different translations. For example, the idiom "Keep at bay" has a different contextual meaning in the following sentences: (i) "The infection is kept at bay." meaning: "The infection is under control." (ii) "The fire keeps the wolves at bay." meaning: "The fire keeps the wolves away."

Recent large language models (LLMs) have enabled improvements in idiom translation compared to NMT models (Raunak et al., 2023), due to their higher tendency towards non-literalness (Raunak et al., 2023) and greater paraphrastic capability (Hendy et al., 2023). However, no research has yet compared the performance of LLMs across different prompts, prompting techniques, and in combination with NMT models.

In this research, focusing on Persian-English translation, we try to fill several gaps. Since Persian datasets are limited in this context, we first introduce a comprehensive resource for idiomatic expressions in Persian (PersianIdioms). This resource captures idiomatic expressions and their meanings, including contextual usage examples. Additionally, we produce two parallel En→Fa and Fa→En datasets, each consisting of 200 sentences containing English and Persian idiomatic expressions. We then use these datasets to evaluate the performance of open-source—Qwen-2.5-72B (Team, 2024) and Command R+ (Cohere For AI, 2024)—, and closed-source—GPT-3.5 (OpenAI, 2023), GPT-4o-mini (OpenAI et al., 2024) and Claude-3.5-Sonnet (Anthropic, 2024)— LLMs, open-source—NLLB-200-3.3b (Team et al., 2022),

* Equal contribution.

[1]Datasets and evaluation guide available at https://github.com/Sara-Rezaeimanesh/Fa-En-Idiom-Translation

MADLAD-400-10b (Kudugunta et al., 2023)— and closed-source— Google Translate— NMT models, and a combination of them in idiom translation across various settings.

We manually assess translation quality using two metrics, idiom translation and fluency. Then, we explore suitable replacements for manual evaluation by calculating the correlation between scores from existing automatic evaluation approaches and manually obtained scores.

In summary, our main contributions are as follows. (i) PersianIdioms, a new resource for Persian idioms which includes about 2,200 idioms and their meanings—a resource that previously did not exist for Persian. A subset of 700 idioms also includes example usages. (ii) Parallel Fa→En and En→Fa datasets each containing 200 examples with at least one idiom. The Persian sentences are sourced from PersianIdioms, while the English sentences are primarily drawn from EPIE (Saxena and Paul, 2020) and MAGPIE (Xu et al., 2024). (iii) A comprehensive comparison of human evaluations versus LLMs-as-a-judge, and existing automatic evaluation methods in assessing translation quality of sentences containing idiomatic expressions. (iv) An evaluation of different prompting methods and a combination of LLMs and NMT models on idiom translation, highlighting their performance differences between Fa→En and En→Fa.

## 2 Related work

### 2.1 Idiom datasets

Xu et al. (2024) present the largest English idiom corpus to date, containing over 50K instances, by using a fixed idiom list, automatic pre-extraction, and a controlled crowdsourced annotation process. Saxena and Paul (2020) compile the EPIE dataset of sentences containing highly occurring English idioms and idioms using StringNet. Kabra et al. (2023) create the MABL dataset covering the figurative language from 7 typologically diverse languages, highlighting cultural and linguistic variations. Liu et al. (2023) investigate the ability of multilingual language models (mLLMs) to reason with cultural common ground by using idioms and sayings as a proxy. They construct a new dataset called MAPS, covering 6 languages with idioms, conversational usages, interpretations, and figurative labels. Li et al. (2024) present a methodology for constructing a large-scale, multilingual idiom knowledge base by distilling figurative meanings from language models. Liu et al. (2022) introduce Fig-QA, a new task to test language models' ability to interpret figurative language. They crowdsource a dataset of over 10k paired metaphorical phrases with opposite meanings and literal interpretations.

These works demonstrate techniques for compiling figurative language data across multiple languages. However, they are focused on English or non-Persian languages, leaving a gap for a large-scale Persian idiom dataset. This research applies similar techniques of utilizing existing resources and language model generation to create idiom data specifically for Persian.

### 2.2 Translation and LLMs

Jiao et al. (2023) demonstrate that ChatGPT competes well with translation services like Google Translate for high-resource European languages but struggles with low-resource or distant languages. Moslem et al. (2023) evaluate the performance of GPT-3.5 and GPT-4' in adaptive MT, comparing it to strong MT systems and show that GPT-3.5 excels in high-resource languages but struggles with low-resource ones, where traditional models perform better. Hendy et al. (2023) suggest that the increased tendency for paraphrasing in GPT translations could assist NMT models in translating figurative language. We validate this hypothesis empirically in our paper in the case of English and Persian translations. Yamada (2024) offers two prompts aimed at enhancing the quality of translations generated by ChatGPT. We assess and contrast these prompts with our approaches. Raunak et al. (2023) propose novel evaluation metrics for measuring translation literalness and compare the performance of GPT models and NMT models in idiom translation, finding that translations produced by GPT models are generally less literal. Several other studies have been dedicated to comparing the performance of different LLMs and NMT models for different languages (Castaldo et al., 2024; Zhu et al., 2024; Tang et al., 2024). Others have aimed to enhance LLM idiom translation through fine-tuning (Stap et al., 2024) and augmenting them with knowledge bases (Li et al., 2024).

However, these efforts primarily focus on individual model performances, overlooking the effects of more complex prompting techniques and the potential advantages of combining these models. Additionally, idiom translation between English and Persian remains underexplored. This work aims to fill these gaps.

## 3 Datasets

### 3.1 PersianIdioms

Our data collection begins with extracting Persian idioms and their meanings from an online dictionary called Abadis[2]. For each idiom, we also gathered usage examples, sourced from user-generated examples in Abadis, to provide contextual clarity. These examples are crucial for future testing of language models, allowing them to learn from idiomatic expressions in use. This comprehensive dataset of Persian idioms, their meanings, and contextual usage examples has never existed before, making it a valuable resource for the development and evaluation of language models for Persian.

**Data verification** Once the dataset is collected, it undergoes a thorough cleaning process. Native Persian speakers review the idioms, verifying the accuracy of their meanings, their cultural relevance, and the appropriateness of the usage examples. The resulting dataset comprises 2,200 idiom-meaning pairs, with 700 idioms enriched with contextual examples. This dataset highlights the richness and complexity of Persian idiomatic expressions and serves as a valuable resource for advancing NLP research in interpreting culturally nuanced language.

### 3.2 Translation datasets

**Fa→En** To ensure representativeness, we sort the idioms containing contextual examples in our PersianIdioms dataset by the number of Google searches and randomly select 200 samples using a uniform distribution. These selected idioms are then manually reviewed to exclude any that are outdated or rarely used. Additionally, we refine the samples to ensure they contain only a single idiom, simplifying complex expressions when necessary. Finally, an expert translator produces the English translations, which are then reviewed and validated by another expert. Table 1 shows an example of our dataset.

**En→Fa** In the initial data collection phase, we attempted to identify sentences containing idiomatic expressions from existing En→Fa parallel resources. However, we found that the Persian translations in these datasets were either automatically generated, derived from translations of English literature into Persian (Kashefi, 2020), or sourced from Wikipedia (Karimi et al., 2019). Each of these approaches poses significant challenges to our research objectives. Automatic translation by NMT models, often produces inaccurate results, especially for figurative language, which is the focus of this study. Literary translations tend to incorporate contextual references, such as character names, or modify sentence structures and meaning to enhance fluency in the target language. Wikipedia-sourced sentences lack complex, culturally specific idioms and primarily feature easily translatable expressions like "under pressure".

| | |
|---|---|
| **Idiom**<br>**Meaning**<br>**Meaning in English** | اب دوغ خیاری<br>پیش پاافتاده / مبتذل<br>low quality/tasteless |
| **Example** | هروقت می رم خونه شون همه پای تلویزیون نشسته ان و دارن یکی از این فیلم های اب دوغ خیاری رو تماشا می کنن. |
| **Gold translation** | Every time I go to their house, everyone is sitting in front of the TV watching one of those low-quality movies. |

Table 1: An example from the Fa→En dataset

Given these limitations, we opt for manual data collection. Drawing primarily from the EPIE and MAGPIE datasets, we carefully select sentences that emphasize the challenges of translating idiomatic expressions, rather than overall sentence structure and exclude outdated idioms. A proficient translator then produces Persian renditions of these selected sentences, followed by a review and validation process conducted by another qualified expert. The result of these efforts is a dataset comprising 200 pairs of English sentences and their Persian translations.

## 4 Methodology

### 4.1 Translation

We use NMT models, LLMs, and a hybrid approach combining LLMs and NMT models to generate translations. The hybrid approach first guides the LLM to identify and replace idioms with literal expressions, and then applies NMT to translate the resulting text into the target language.

The prompts used for LLMs in En→Fa translation are shown in Table 2. The second single prompt is taken from the prompts presented in Yamada (2024). Fa→En prompts replace "English" with "Persian" and vice versa, and "American" with "Iranian". Accordingly, we consider five prompts grouped into three categories: (i) SinglePrompt: three single prompts (ii) CoTPrompt: one chain of thought (CoT) prompt, and (iii) MultiPrompt: Multiple prompts that break down a single prompt into multiple independent steps. In the CoT setup, the three steps are provided as separate prompts, with each step and the model's response carried forward as chat history for the subsequent step. In contrast, the third category eliminates the reliance on chat history by using two independent prompts, where the answer to the first prompt is embedded within the second prompt itself. Initially, we experimented with a single prompt encompassing all three CoT steps. However, manual evaluation revealed that breaking the process into multiple prompts significantly improves the model's ability to follow the instructions accurately.

[2] https://abadis.ir/ (The Abadis website mentions that using the entries of this dictionary is permitted, if the resource is cited.)

| | |
|---|---|
| SinglePrompt | Translate this sentence to Persian. |
| | Translate the following English text into Persian. Use natural expressions that can be understood by Persian speakers, unfamiliar with American Culture. |
| | Translate the following English text into Persian. Avoid word-for-word translations. |
| CoTPrompt | 1) Identify the idioms in this sentence. 2) Replace the idioms with literal clauses. 3) Translate the literal sentence to Persian. Avoid word-for-word translation. |
| MultiPrompt | Identify the idioms in this sentence and replace them with literal clauses. 2) Translate this literal sentence to Persian. Avoid word-for-word translation. |

Table 2: LLM Prompts used for En→Fa translation.

### 4.2 Manual evaluation

Using the MQM evaluation framework (Lommel et al., 2014), we devise two independent evaluation metrics: **fluency** and **idiom translation**. Idiom translation, a binary metric (0 or 1), assesses whether the translation preserves the idiom's meaning within the context of the sentence. Fluency, rated from 1 to 5, evaluates the syntactic and semantic correctness of the translation, assuming the idiom is correctly translated. Idiom translation focuses solely on semantic accuracy, with any grammatical errors in idiom translation affecting only the fluency score and not the idiom translation score.

We distill adequacy down to idiom translation for two reasons. First, our dataset consists of single sentences, that shift the translation challenge to the idiom itself. Therefore, the semantic accuracy of the entire sentence usually depends on the semantic accuracy of the idiom translation. Second, idioms are the core of this study, and we aim to improve idiom translation without compromising overall performance. Thus, occasional non-idiom-related semantic errors only affect fluency scores.

We chose binary labels over a 1-5 scale for idiom translation since idioms, being short phrases, rarely have partially correct translations. While a 1-5 scale might reflect how closely a translation aligns with the intended meaning, it is subjective and heavily influenced by factors like cross-linguistic transferability and the reader's interpretation. Meanwhile, the binary label simply checks whether the meaning of the idiom is correctly conveyed, which simplifies the evaluation process and makes the assessment more objective. If the translation of the idiom preserves its meaning but sounds unnatural, it is treated as a fluency issue, not an idiom translation error.

### 4.3 Automatic evaluation

Manual evaluation is labor-intensive and time-consuming, making automation a valuable step toward streamlining idiom translation research. We experiment with several standard automatic metrics and methods, as well as LLM as judges, and calculate Spearman's correlation between manual and automatic scores. The automatic metrics with the highest correlation are chosen as the best fits for ranking idiom translation performance and fluency of model outputs.

**Existing automatic evaluation metrics** We use BLEU (Papineni et al., 2002), BERTScore (Zhang et al., 2020), and COMET (Rei et al., 2020) as standard evaluation metrics for translation tasks.

**GPT-4o** We follow the LLM-as-a-judge trend, using single-answer and reference-guided grading as in (Zheng et al., 2023; Li et al., 2024) and GPT-4o. For En→Fa, we used the prompt "Is the idiom in this sentence correctly translated into English/Persian? Answer with just a number: 1 for yes and 0 for no. idiom: <idiom>, sentence: <reference>', translation: <model translation>". For translating from Persian, a lower resource language, including a gold translation as a reference, improves correlation, helping the model better assess the accuracy of idiom translations by providing additional guidance. However, for En→Fa, GPT-4o assigns a score of 1 only to translations closely resembling the gold standard, leading to false negatives due to the flexibility of idiom rewrites. We also provide three examples to emphasize the importance of accurate idiom translation and set the temperature hyperparameter to 0.1 to minimize response variations.

## 5 Experimental setup and results

### 5.1 Translation models and prompts

We generate translations using various open- and closed-source NMT models (NLLB-200-3.3b, MADLAD-400-10b, Google Translate[3]), LLMs (GPT-3.5-turbo, QWEN-2.5-72b, Command R+-104b, GPT-4o-mini, Claude-3.5-Sonnet), and the combination of these LLMs and NMT models.[4].

For LLMs, we set the temperature to 0.8 to reduce response variability while preserving some creative freedom. To prevent this variation from skewing evaluations, we ran our experiments with GPT models multiple times. Although individual sentence scores fluctuated between runs, the overall score remained consistent or changed only slightly, confirming that response variation does not significantly affect the final evaluation results.

### 5.2 Manual evaluation results

We compute inter-annotator agreement for the metrics introduced in Section 4.2. Three native Persian-speaking MSc students (some of them are the authors), fluent in English, were given detailed evaluation guidelines with examples and tasked with manually scoring the first 100 sentences from seven outputs generated by GPT-3.5, Google Translate, and their combination. The GPT-3.5 outputs are produced using the prompts outlined in Section 4.1. Idiom translation labels are decided by majority vote, and fluency scores are averaged across annotators' ratings. We also assess inter-annotator agreement for both idiom translation and fluency to ensure reliability.

Fluency scores are highly skewed, with most labels falling between 3 and 5, and the 1–5 scale being inherently subjective. Consequently, metrics like Fleiss' Kappa may overestimate chance agreement, leading to a pessimistic assessment of inter-annotator agreement. To mitigate this, we use Gwet's AC1 (Gwet, 2008), which is less sensitive to label prevalence and better suited for subjective tasks. In contrast, idiom translation labels are more objective and clearly defined, making Fleiss' Kappa appropriate for assessing inter-annotator agreement. We also report observed fluency agreement based on a 1-point difference threshold and observed idiom translation agreement based on the proportion of sentences with matching idiom translation labels.

Table 3 shows the inter-annotator agreement scores. For idiom translation, annotators align well in both directions, with Kappa exceeding 0.6 and high observed agreement. Fluency scores show moderate agreement based on Gwet's AC1, with a slightly higher score for Persian. The high observed accuracy further supports the reliability of these ratings. Given the task's subjectivity and complexity, these fluency agreement levels are acceptable. (Castilho, 2020) also shows that for fluency, inter-annotator agreement tends to be slight to fair.

| Models | Agreement En-Fa | Agreement Fa-En |
|---|---|---|
| Fluency (Gwen-AC1) | .45 | .54 |
| Fluency (Observed) | .84 | .83 |
| Idiom Translation (Fleiss Kappa) | .63 | .68 |
| Idiom Translation (Observed) | .74 | .73 |

Table 3: Inter-annotator agreement for fluency and idiom-translation

### 5.3 Reliability of automatic evaluation metrics

Table 4 shows the correlations between manual and automatic evaluation scores for the seven manually evaluated model outputs. To focus on overall model performance, correlations are computed using the aggregated scores of the first 100 sentences from each model output (e.g., average fluency and the percentage of correctly translated idioms), rather than individual sentence scores. Consequently, each metric produces a list of seven aggregate scores, corresponding to the seven manually evaluated outputs. As an example, the following two arrays contain the idiom translation and GPT-4o scores of seven different model outputs for Fa→En: Idiom Translation = [.36, .26, .22, .31, .41, .43, .52] and GPT-4o = [.34, .17, .33, .36, .53, .51, .46]. We report Spearman's correlation of these two arrays as the correlation between idiom translation and GPT-4o scores in Table 4 (i.e., 0.79).

Although the sample size for correlations is small, using aggregated scores from various setups reduces noise and highlights meaningful trends.

[3]We use the term NMT to refer to systems specializing in Machine Translation in general as opposed to general purpose models like LLMs.

[4]We utilized the OpenAI API and https://openrouter.ai/ to access these models, incurring an approximate cost of $60 in total for API usage.

The metrics also show consistent results across both translation directions, underscoring their reliability.

Based on Table 4, GPT-4o scores show the highest correlation with idiom translation in both directions, highlighting the model's strong grasp of idioms in Persian and English. As expected, the correlation is higher for English→Persian, indicating that GPT-4o is more adept at identifying and interpreting English idioms than Persian ones.

In both translation directions, BLEU penalizes non-literal translations, while COMET exhibits a stronger correlation with idiom translation compared to BLEU and BERTScore. This is likely due to its consideration of both the source sentence and gold translation, which reduces its correlation with fluency, especially in En→Fa. A more in-depth analysis of these metrics and their behavior is deferred to future work. Similar to human evaluations for fluency, BLEU and BERTScore favor Google Translate, despite its tendency to translate most idioms literally. Given that idioms are usually brief phrases that constitute a small part of the sentence, minimal paraphrasing often yields higher BLEU and BERT scores, as the majority of the produced translation remains closer to the gold reference. Furthermore, correctly translated idioms might still differ from the gold translation and fail to score higher than literal translations. This explains why fluency, a metric independent of idiom translation, continues to exhibit a high correlation with these other metrics.

Interestingly, the correlation between BLEU and fluency for Fa→En translations is lower compared to En→Fa. This suggests that Fa→En translation may involve more paraphrasing and structural changes. However, these correlations are not strong enough to draw definitive conclusions.

Ultimately, the correlations show that for En→Fa, BLEU, and GPT-4o, and for Fa→En, BERTScore, and GPT-4o are well-suited for ranking fluency and idiom translation performance of model outputs, respectively.

We further examine GPT-4o's performance as a judge by calculating the agreement percentage between manually obtained idiom translation scores and GPT-4o labels. Table 5 compares the average agreement between human annotator pairs and between GPT-4o and each annotator. The agreement between GPT-4o and human annotators approaches the average inter-annotator agreement, suggesting that GPT-4o performs comparably to humans and can serve as a reliable evaluation tool for idiom translation. Our manual inspections show that GPT-4o tends to slightly underestimate model performance, sometimes labeling idiom translations as incorrect when they involve significant paraphrasing or due to errors elsewhere in the sentence. Nonetheless, as the correlations and agreements indicate, it remains a reliable tool for ranking model performance in idiom translation.

| | Metric | COMET | BERTScore | BLEU | GPT-4o |
|---|---|---|---|---|---|
| En→Fa | Fluency | .17 | .89 | **.96** | -.35 |
| | IdiomT | .63 | .18 | -.03 | **.88** |
| Fa→En | Fluency | .72 | **.88** | .67 | .15 |
| | IdiomT | .53 | .25 | -.03 | **.79** |

Table 4: Spearman's Correlation between results obtained from automatic and manual evaluation for En→FA and Fa→En. The best correlation for each row is in bold. IdiomT:idiomtranslation

| | | Human | GPT-4o |
|---|---|---|---|
| En→Fa | Human | .81 | .76 |
| Fa→En | Human | .73 | .71 |

Table 5: Average agreement % on idiom translation between human annotators and GPT-4o.

### 5.4 Comprehensive evaluation: En→Fa

The left side of Table 6 presents the results for En→Fa translation. As discussed in Section 5.3 BLEU and GPT-4o scores are the most suitable metrics for ranking model performance in terms of fluency and idiom translation in this direction and will be the primary focus of this section. It is important to note that n-gram-based metrics like BLEU are ill-suited for figurative language since they prioritize exact matches over semantic similarity, explaining the low BLEU scores even when other metrics indicate better performance. Nevertheless, BLEU remains a useful metric for **ranking** model performance based on fluency, as shown by its Spearman correlation.

**Best models** Claude-3.5-Sonnet achieves the highest BLEU in the SinglePrompt and the highest GPT-4o score in the CoTPrompt setup, making it the most effective model overall. Our manual inspections reveal that this model not only excels in accurately identifying and understanding idioms but also finds suitable Persian idiom replacements, contributing to its strong translation capabilities.

| | En→Fa | | | | Fa→En | | | |
|---|---|---|---|---|---|---|---|---|
| | COMET | BERTScore | BLEU | GPT4o | COMET | BERTScore | BLEU | GPT4o |
| GPT-3.5 | | | | | | | | |
| • SinglePrompt1 | 82.6 | 82.3 | 11.9 | 63.5 | **75.3** | **93.5** | 23.2 | 36.0 |
| • SinglePrompt2 | 82.7 | 81.6 | 8.6 | 65.0 | 74.8 | 93.1 | 20.6 | 40.0 |
| • SinglePrompt3 | **83.0** | 82.3 | 11.1 | 65.5 | 74.8 | 93.1 | 21.8 | **43.0** |
| • CoTPrompt | 82.1 | 81.2 | 8.3 | 68.0 | 72.4 | 92.6 | 19.3 | 29.0 |
| • MultiPrompt | 81.4 | 80.7 | 7.9 | 72.0 | 71.6 | 92.6 | 18.4 | 30.0 |
| • +GT | 85.1 | **84.6** | **19.3** | **79.0** | 74.5 | 92.8 | **25.3** | 25.0 |
| • +NLLB | 81.5 | 78.8 | 9.3 | 64.0 | 73.1 | 92.6 | 20.9 | 26.0 |
| • +Madlad | 80.8 | 77.2 | 10.0 | 63.5 | 73.7 | 92.6 | 21.7 | 25.0 |
| Qwen 2.5 72B | | | | | | | | |
| • SinglePrompt1 | 82.6 | 84.3 | 14.7 | 66.0 | 75.3 | 93.5 | 26.4 | 35.0 |
| • SinglePrompt2 | 83.0 | 82.1 | 13.4 | 72.0 | 76.5 | 93.4 | 23.8 | **41.5** |
| • SinglePrompt3 | 83.0 | 81.8 | 12.2 | 74.5 | 76.5 | 93.6 | 25.4 | 39.5 |
| • CoTPrompt | 80.0 | 79.5 | 7.2 | 72.5 | 75.8 | 93.6 | **27.2** | 34.5 |
| • MultiPrompt | 80.5 | 80.1 | 9.2 | 74.0 | **76.7** | **93.7** | 26.5 | 35.5 |
| • +GT | **84.2** | **83.7** | **17.9** | **88.0** | 74.0 | 93.0 | 24.6 | 30.0 |
| • +NLLB | 81.2 | 78.9 | 8.5 | 65.5 | 72.2 | 92.3 | 19.3 | 24.0 |
| • +Madlad | 80.7 | 77.3 | 10.0 | 70.5 | 70.1 | 91.8 | 17.5 | 16.5 |
| GPT4o-mini | | | | | | | | |
| • SinglePrompt1 | 85.0 | 84.5 | 18.7 | 85.0 | 77.16 | 94.2 | 29.6 | 52.0 |
| • SinglePrompt2 | 85.5 | 84.5 | **19.9** | 90.0 | 79.4 | **94.4** | 27.0 | 56.0 |
| • SinglePrompt3 | **85.8** | **84.6** | 16.5 | 87.5 | **79.6** | 94.3 | 26.3 | **62.0** |
| • CoTPrompt | 84.5 | 83.2 | 15.1 | 90.5 | 79 | 93.9 | 25.0 | 52.5 |
| • MultiPrompt | 83.9 | 83.1 | 15.7 | **91.0** | 78.9 | 94.0 | **30.0** | 55.5 |
| • +GT | 84.7 | 84.0 | 17.3 | 87.0 | 79.0 | 93.7 | 26.9 | 54.0 |
| • +NLLB | 80.8 | 78.4 | 8.5 | 64.0 | 75.5 | 92.7 | 18.9 | 36.0 |
| • +Madlad | 80.4 | 76.6 | 8.5 | 65.0 | 73.5 | 92.2 | 17.1 | 31.0 |
| Command R+ | | | | | | | | |
| • SinglePrompt1 | 83.2 | 82.2 | 12.2 | 78.5 | 75.0 | 93.0 | 21.1 | 55.5 |
| • SinglePrompt2 | 82.6 | 81.5 | 10.5 | 75.0 | 76.7 | 93.2 | 21.0 | 57.0 |
| • SinglePrompt3 | 83.3 | 81.6 | 10.1 | **83.5** | 75.8 | 92.2 | 14.7 | 52.5 |
| • CoTPrompt | 78.7 | 78.8 | 6.9 | 69.5 | 70.9 | 91.3 | 13.8 | **60.5** |
| • MultiPrompt | 79.3 | 78.5 | 5.9 | 70.0 | 74.5 | 92.3 | 15.1 | 55.5 |
| • +GT | **84.4** | **83.9** | **17.7** | 81.5 | **77.4** | **93.3** | **23.6** | 57.0 |
| • +NLLB | 80.6 | 78.4 | 8.3 | 56.0 | 75.1 | 92.5 | 18.2 | 37.0 |
| • +Madlad | 80.0 | 76.4 | 9.4 | 66.0 | 72.6 | 92.0 | 16.1 | 30.5 |
| Claude 3.5 Sonnet | | | | | | | | |
| • SinglePrompt1 | 85.1 | **84.6** | **21.1** | 91.0 | 79.7 | 94.6 | **32.1** | 68.0 |
| • SinglePrompt2 | 85.6 | 84.5 | 19.9 | 93.0 | 78.9 | 94.3 | 25.9 | 71.0 |
| • SinglePrompt3 | **86.0** | 84.4 | 20.8 | 93.5 | 79.2 | 94.3 | 24.7 | 70.5 |
| • CoTPrompt | 84.3 | 82.9 | 15.3 | **94.0** | 82.1 | 94.4 | 24.2 | 74.0 |
| • MultiPrompt | 83.5 | 82.9 | 17.0 | 92.5 | **82.8** | **94.8** | 29.8 | **75.0** |
| • +GT | 84.7 | 84.2 | 18.5 | 92.0 | 77.4 | 93.3 | 23.6 | 61.0 |
| • +NLLB | 81.4 | 79.3 | 9.9 | 70.0 | 77.6 | 93.3 | 21.3 | 50.5 |
| • +Madlad | 81.2 | 77.9 | 10.7 | 72.5 | 76.4 | 93.1 | 20.3 | 45.5 |
| Google Translate (GT) | | | | | | | | |
| • Direct Translation | 81.1 | 83.7 | 17.6 | 52.0 | 73.9 | 93.1 | 26.1 | 21.0 |
| NLLB-200-3.3b | | | | | | | | |
| • Direct Translation | 77.4 | 77.0 | 7.3 | 34.0 | 70.5 | 92.2 | 19.3 | 18.0 |
| MADLAD-400-10b | | | | | | | | |
| • Direct Translation | 78.1 | 75.2 | 8.6 | 54.5 | 72.2 | 92.4 | 19.4 | 22.0 |

Table 6: Automatic evaluation results for different models and setups tested on 200 samples. For each translation direction, the highest scores for each metric across all models are underlined, while the highest scores within each model are in bold.

**Prompt performance** Across all models in the SinglePrompt setup, the second and third prompts consistently outperform the first. The first prompt simply requests a sentence translation, increasing the likelihood of literal idiom translations. In contrast, the second and third prompts provide more guidance, reducing literal translations but slightly compromising fluency.

| | En →Fa Translation | Fa →En Translation |
|---|---|---|
| **Sentence** | Poor Mrs has lots of children and they were driving her up the wall! | من اعتماد کردم و حرف دلم را به او زدم اما او به هر کس رسید همه را روی دایره ریخت |
| **Gold Translation** | خانم بیچاره بچه های زیادی دارد که او را آزرده خاطر میکردند! | I trusted him and spoke out my heart, but he **revealed all my secrets to everyone he encountered.** |
| **Google Translate** | خانم بیچاره بچه های زیادی دارد که او را از دیوار بالا میبرند.<br>Poor Mrs has lots of children that take her up the wall. | I trusted and spoke my heart to her, but she **threw everyone on the circle.** |
| **Single Prompt** | خانم فلانی بچه های زیادی دارد که دارند به اعصابش میزنند.<br>Some Mrs has lots of children that are hitting her nerves. | I trusted him and opened up to him, but he **betrayed my trust by sharing everything with everyone.** |
| **Chain Prompt** | مادر بدبخت بسیاری فرزند داشت و آن‌ها باعث ایجاد تنش شدید برای او بودند.<br>Poor mother had many children that had been causing her a lot of tension. | I trusted him and confided in him, but he **betrayed everyone and turned against all.** |
| **Multi Prompt** | خانم فقیر دارای بسیاری فرزند است و آن‌ها باعث ایجاد تنش شدید برایش می‌شوند!<br>Poor (in Persian, the word used means impoverished, not unfortunate) Mrs has many children that have been causing her extreme tension! | I trusted him and spoke my heart out to him, but he **reached out to everyone and spread it on everyone's face.** |
| **GPT-3.5 + Google Translate** | خانم بیچاره بچه های زیادی دارد که او را اذیت می کردند!<br>Poor mother had many children that are bothering her! | I trusted and spoke out my heart to her, but she **reached out to everyone and spread everything to their face.** |

Table 7: Examples of GPT-3.5 En →Fa and Fa →En translations with back-translations to show translation quality.

For most models, the CoT or MultiPrompt setup, which breaks the task into smaller steps, improves GPT-4o scores but hurts fluency. Manual inspections reveal that when given more complicated prompts, models tend to paraphrase more, over-explain idiom definitions, add unnecessary content to the sentence, or even alter correct idiom definitions into literal ones when provided with additional steps, especially for idioms they could correctly translate without guidance. These issues can lead to deviations from the original sentence, ultimately lowering BLEU scores. In the case of Command R+, GPT-4o scores are also lower due to its frequent over-explanations and additions to both the sentences and idiom definitions.

**The hybrid approach** Google Translate achieves higher BLEU scores than GPT-3.5, Qwen-2.5, and Command R+ across all prompts. When combined with Google Translate, these models show an increased BLEU score, benefiting from the strengths of both LLMs and NMT models. Notably, the GPT-4o score improves significantly for Qwen-2.5 and GPT-3.5 as well. Manual inspections reveal that Qwen-2.5 often uses characters and words from other languages such as Chinese when translating to Persian, which can render the translation meaningless and GPT-3.5 translations suffer from fluency issues that sometimes cause GPT-4o to label them as incorrect. For Command R+, the hybrid model's GPT-4o score nearly matches the model's best, while maintaining superior fluency. In contrast, GPT-4o-mini and Claude-3.5-Sonnet outperform Google Translate in BLEU with certain prompts and, therefore, experience a performance decline when combined with it. NLLB and MADLAD exhibit weaker performance than Google Translate. In general, their translations are less fluent, with lower fidelity to the original sentence and occasional alterations. Moreover, their fluency is lower than that of LLMs, and combining them with LLMs leads to decreased GPT-4o and BLEU scores for all models.

A key takeaway from these results is that weaker models like Qwen-2.5-72b, when combined with NMT models that exhibit a higher fluency score, can perform comparably to much stronger models such as Claude-3.5-Sonnet. This suggests that when an LLM's fluency is lower than that of an NMT model, combining the two can effectively close the performance gap with stronger LLMs.

**An example** the left side of Table 7 highlights the challenges in En→Fa translation for GPT-3.5 through an example. Google Translate offers fluent translations but often renders idioms literally. GPT-3.5 correctly detects idioms but produces unnatural definitions, producing more fluent translations with CoT and MultiPrompt. Combining GPT-3.5 with Google Translate yields the most fluent translations.

### 5.5 Comprehensive evaluation: Fa→En

The right side of Table 6 shows the automatic evaluation results for Fa→En translation. As discussed in Section 5.3, BERTScore and GPT-4o are the most appropriate metrics for ranking model performance in fluency and idiom translation in this direction. Therefore, we focus primarily on these two metrics.

**Best models** GPT4o-mini and Claude-3.5 Sonnet excel in idiom translation, outperforming other models across all metrics. GPT4o-mini with SinglePrompt 3 and Claude-3.5 Sonnet with MultiPrompt deliver the most accurate, contextually aware, and fluent translations. Claude-3.5-Sonnet frequently selects appropriate English idioms as replacements, demonstrating a strong understanding of both Persian and English idioms.

**Prompt performance** Like En→Fa, in the SinglePrompt setup, the second and third prompts improve GPT-4o scores across all models but slightly sacrifice fluency. GPT-3.5, GPT-4o-mini, and Qwen-2.5 perform best with single prompts and struggle with more complex setups like MultiPrompt and CoTPrompt, frequently failing to accurately identify or translate idioms within the provided context. Manual inspection of the GPT-3.5 outputs reveals that, in these setups, the model often identifies idioms, removes them from the sentence, and translates them outside the given context. This loss of context reduces its idiom translation performance, especially since the model is not well-versed in Persian idioms. Similar behavior is observed with Qwen-2.5 and GPT-4o-mini. However, when single prompts are used, these LLMs are more likely to produce accurate idiom translations by leveraging sentence context. In contrast, Claude-3.5-Sonnet and Command R+ achieve their highest GPT-4o scores using complex prompting setups. Claude-3.5-Sonnet outputs show that even in these setups, sentence context is considered during idiom translation. For Command R+, the CoTPrompt approach aids in better idiom detection, and the model often does not detect idioms and translates them literally in the SinglePrompt setup.

Finally, in the CoT and MultiPrompt setups, idioms are replaced with their meanings in **Persian** sentences, which might disrupt sentence structure and introduce additional fluency issues since LLMs are often not fluent enough in Persian to make the necessary adjustments after idiom replacement.

**The hybrid approach** In Fa→En, all NMT models fall behind in idiom translation and fluency. NLLB often omits parts of the sentences in translation and MADLAD replaces unknown words (e.g. names) with random characters. Google Translate demonstrates a comparable performance to LLMs and when their fluency drops below that of Google Translate in the CoT or MultiPrompt setup, delegating the translation step to Google Translate can enhance fluency, as observed in the cases of GPT-3.5 and Command R+. However, this approach often hurts the GPT-4o score. Combined with the observation that most models achieve higher BERTScores independently, this suggests that in general, LLMs perform better in translating Persian sentences with idiomatic expressions.

**An example** The left side of Table 7 highlights the challenges in Fa→En translations through an example. Google Translate often translates idioms literally. GPT-3.5 struggles with complex prompts, occasionally misidentifying or misinterpreting idioms, and performs best with a single prompt. Although combining GPT-3.5 with Google Translate yields more fluent results, it does not demonstrate superior idiom translation performance.

### 5.6 Comparing Fa→En and En→Fa results

GPT-4o scores are significantly higher for En→Fa translation, with even the strongest-performing model for Fa→En showing poorer performance than the weakest model in En→Fa. This highlights that models are far more familiar with English idioms than Persian ones, emphasizing the challenges of idiom translation in lower-resource languages. Moreover, all LLMs show higher BLEU and BERT scores for Fa→En, likely due to the models' stronger understanding of English which makes them more proficient at producing English sentences rather than Persian ones.

## 6 Conclusion

We introduced two parallel datasets for Fa→En and En→Fa idiom translation. The Persian idioms were sampled from our PersianIdioms resource, with 2,200 idioms and their meanings. Using these datasets, we evaluated multiple LLMs, NMT models, and their combination, focusing on idiom translation accuracy and fluency. Our results show that Claude-3.5-Sonnet performs best in both directions. We also found that models generally translate English idioms more effectively than Persian ones. Performance varies by translation direction—for En→Fa, combining weaker models with Google Translate improves their performance, and for Fa→En, weaker models performed best with single prompts. Stronger models performed best with CoT or multiple prompts in both directions. Additionally, we evaluated existing automatic metrics and GPT-4o as a judge, confirming GPT-4o's reliability for assessing idiom translation accuracy.

## 7 Limitations

Our work is limited in several aspects, which we briefly discuss here.

- Our parallel datasets contain only 200 examples for each translation direction. Expanding its size could enhance both the quality of the data set and the robustness of our findings.
- We focus only on Persian and English. Extending to other languages would help determine whether some of our observations are generalizable or not.

## References

Anthropic. 2024. Claude 3.5 sonnet model card addendum.

Antonio Castaldo, Johanna Monti, et al. 2024. Prompting large language models for idiomatic translation. In *Proceedings of the First Workshop on Creative-text Translation and Technology*, pages 37–44.

Sheila Castilho. 2020. On the same page? comparing inter-annotator agreement in sentence and document level human machine translation evaluation. In *Proceedings of the Fifth Conference on Machine Translation*, pages 1150–1159, Online. Association for Computational Linguistics.

Cohere For AI. 2024. c4ai-command-r-plus (revision 432fac1).

Verna Dankers, Christopher Lucas, and Ivan Titov. 2022. Can transformer be too compositional? analysing idiom processing in neural machine translation. In *Proceedings of the 60th Annual Meeting of the Association for Computational Linguistics (Volume 1: Long Papers)*, pages 3608–3626, Dublin, Ireland. Association for Computational Linguistics.

Kilem Li Gwet. 2008. Computing inter-rater reliability and its variance in the presence of high agreement. *British Journal of Mathematical and Statistical Psychology*, 61(1):29–48.

Amr Hendy, Mohamed Abdelrehim, Amr Sharaf, Vikas Raunak, Mohamed Gabr, Hitokazu Matsushita, Young Jin Kim, Mohamed Afify, and Hany Hassan Awadalla. 2023. How good are gpt models at machine translation? a comprehensive evaluation. *Preprint*, arXiv:2302.09210.

Wenxiang Jiao, Wenxuan Wang, Jen tse Huang, Xing Wang, Shuming Shi, and Zhaopeng Tu. 2023. Is chatgpt a good translator? yes with gpt-4 as the engine. *Preprint*, arXiv:2301.08745.

Anubha Kabra, Emmy Liu, Simran Khanuja, Alham Fikri Aji, Genta Winata, Samuel Cahyawijaya, Anuoluwapo Aremu, Perez Ogayo, and Graham Neubig. 2023. Multi-lingual and multi-cultural figurative language understanding. In *Findings of the Association for Computational Linguistics: ACL 2023*, pages 8269–8284, Toronto, Canada. Association for Computational Linguistics.

Akbar Karimi, Ebrahim Ansari, and Bahram Sadeghi Bigham. 2019. Extracting an english-persian parallel corpus from comparable corpora. *Preprint*, arXiv:1711.00681.

Omid Kashefi. 2020. Mizan: A large persian-english parallel corpus. *Preprint*, arXiv:1801.02107.

Sneha Kudugunta, Isaac Caswell, Biao Zhang, Xavier Garcia, Christopher A. Choquette-Choo, Katherine Lee, Derrick Xin, Aditya Kusupati, Romi Stella, Ankur Bapna, and Orhan Firat. 2023. Madlad-400: A multilingual and document-level large audited dataset. *Preprint*, arXiv:2309.04662.

Shuang Li, Jiangjie Chen, Siyu Yuan, Xinyi Wu, Hao Yang, Shimin Tao, and Yanghua Xiao. 2024. Translate meanings, not just words: Idiomkb's role in optimizing idiomatic translation with language models. In *Proceedings of the AAAI Conference on Artificial Intelligence*, volume 38, pages 18554–18563.

Chen Cecilia Liu, Fajri Koto, Timothy Baldwin, and Iryna Gurevych. 2023. Are multilingual llms culturally-diverse reasoners? an investigation into multicultural proverbs and sayings. *arXiv preprint arXiv:2309.08591*. Cs.CL.

Emmy Liu, Chenxuan Cui, Kenneth Zheng, and Graham Neubig. 2022. Testing the ability of language models to interpret figurative language. In *Proceedings of the 2022 Conference of the North American Chapter of the Association for Computational Linguistics: Human Language Technologies*, volume NAACL, pages 4437–4452, Seattle, United States. Association for Computational Linguistics.

Arle Lommel, Hans Uszkoreit, and Aljoscha Burchardt. 2014. Multidimensional quality metrics (mqm): A framework for declaring and describing translation quality metrics.

Yasmin Moslem, Rejwanul Haque, John D. Kelleher, and Andy Way. 2023. Adaptive machine translation with large language models. *Preprint*, arXiv:2301.13294.

Makoto Nagao. 1984. A framework of a mechanical translation between japanese and english by analogy principle. In *Proc. of the International NATO Symposium on Artificial and Human Intelligence*, page 173–180, USA. Elsevier North-Holland, Inc.

OpenAI. 2023. Introducing chatgpt. Available: https://openai.com/blog/chatgpt.

OpenAI, Josh Achiam, Steven Adler, Sandhini Agarwal, Lama Ahmad, Ilge Akkaya, Florencia Leoni Aleman, Diogo Almeida, Janko Altenschmidt, Sam Altman, Shyamal Anadkat, Red Avila, Igor Babuschkin, Suchir Balaji, Valerie Balcom, Paul Baltescu, Haiming Bao, Mohammad Bavarian, Jeff Belgum, Irwan Bello, Jake Berdine, Gabriel Bernadett-Shapiro, Christopher Berner, Lenny Bogdonoff, Oleg Boiko, Madelaine Boyd, Anna-Luisa Brakman, Greg Brockman, Tim Brooks, Miles Brundage, Kevin Button, Trevor Cai, Rosie Campbell, Andrew Cann, Brittany Carey, Chelsea Carlson, Rory Carmichael, Brooke Chan, Che Chang, Fotis Chantzis, Derek Chen, Sully Chen, Ruby Chen, Jason Chen, Mark Chen, Ben Chess, Chester Cho, Casey Chu, Hyung Won Chung, Dave Cummings, Jeremiah Currier, Yunxing Dai, Cory Decareaux, Thomas Degry, Noah Deutsch, Damien Deville, Arka Dhar, David Dohan, Steve Dowling, Sheila Dunning, Adrien Ecoffet, Atty Eleti, Tyna Eloundou, David Farhi, Liam Fedus, Niko Felix, Simón Posada Fishman, Juston Forte, Isabella Fulford, Leo Gao, Elie Georges, Christian Gibson, Vik Goel, Tarun Gogineni, Gabriel Goh, Rapha Gontijo-Lopes, Jonathan Gordon, Morgan Grafstein, Scott Gray, Ryan Greene, Joshua Gross, Shixiang Shane Gu, Yufei Guo, Chris Hallacy, Jesse Han, Jeff Harris, Yuchen He, Mike Heaton, Johannes Heidecke, Chris Hesse, Alan Hickey, Wade Hickey, Peter Hoeschele, Brandon Houghton, Kenny Hsu, Shengli Hu, Xin Hu, Joost Huizinga, Shantanu Jain, Shawn Jain, Joanne Jang, Angela Jiang, Roger Jiang, Haozhun Jin, Denny Jin, Shino Jomoto, Billie Jonn, Heewoo Jun, Tomer Kaftan, Łukasz Kaiser, Ali Kamali, Ingmar Kanitscheider, Nitish Shirish Keskar, Tabarak Khan, Logan Kilpatrick, Jong Wook Kim, Christina Kim, Yongjik Kim, Jan Hendrik Kirchner, Jamie Kiros, Matt Knight, Daniel Kokotajlo, Łukasz Kondraciuk, Andrew Kondrich, Aris Konstantinidis, Kyle Kosic, Gretchen Krueger, Vishal Kuo, Michael Lampe, Ikai Lan, Teddy Lee, Jan Leike, Jade Leung, Daniel Levy, Chak Ming Li, Rachel Lim, Molly Lin, Stephanie Lin, Mateusz Litwin, Theresa Lopez, Ryan Lowe, Patricia Lue, Anna Makanju, Kim Malfacini, Sam Manning, Todor Markov, Yaniv Markovski, Bianca Martin, Katie Mayer, Andrew Mayne, Bob McGrew, Scott Mayer McKinney, Christine McLeavey, Paul McMillan, Jake McNeil, David Medina, Aalok Mehta, Jacob Menick, Luke Metz, Andrey Mishchenko, Pamela Mishkin, Vinnie Monaco, Evan Morikawa, Daniel Mossing, Tong Mu, Mira Murati, Oleg Murk, David Mély, Ashvin Nair, Reiichiro Nakano, Rajeev Nayak, Arvind Neelakantan, Richard Ngo, Hyeonwoo Noh, Long Ouyang, Cullen O'Keefe, Jakub Pachocki, Alex Paino, Joe Palermo, Ashley Pantuliano, Giambattista Parascandolo, Joel Parish, Emy Parparita, Alex Passos, Mikhail Pavlov, Andrew Peng, Adam Perelman, Filipe de Avila Belbute Peres, Michael Petrov, Henrique Ponde de Oliveira Pinto, Michael, Pokorny, Michelle Pokrass, Vitchyr H. Pong, Tolly Powell, Alethea Power, Boris Power, Elizabeth Proehl, Raul Puri, Alec Radford, Jack Rae, Aditya Ramesh, Cameron Raymond, Francis Real, Kendra Rimbach, Carl Ross, Bob Rotsted, Henri Roussez, Nick Ryder, Mario Saltarelli, Ted Sanders, Shibani Santurkar, Girish Sastry, Heather Schmidt, David Schnurr, John Schulman, Daniel Selsam, Kyla Sheppard, Toki Sherbakov, Jessica Shieh, Sarah Shoker, Pranav Shyam, Szymon Sidor, Eric Sigler, Maddie Simens, Jordan Sitkin, Katarina Slama, Ian Sohl, Benjamin Sokolowsky, Yang Song, Natalie Staudacher, Felipe Petroski Such, Natalie Summers, Ilya Sutskever, Jie Tang, Nikolas Tezak, Madeleine B. Thompson, Phil Tillet, Amin Tootoonchian, Elizabeth Tseng, Preston Tuggle, Nick Turley, Jerry Tworek, Juan Felipe Cerón Uribe, Andrea Vallone, Arun Vijayvergiya, Chelsea Voss, Carroll Wainwright, Justin Jay Wang, Alvin Wang, Ben Wang, Jonathan Ward, Jason Wei, CJ Weinmann, Akila Welihinda, Peter Welinder, Jiayi Weng, Lilian Weng, Matt Wiethoff, Dave Willner, Clemens Winter, Samuel Wolrich, Hannah Wong, Lauren Workman, Sherwin Wu, Jeff Wu, Michael Wu, Kai Xiao, Tao Xu, Sarah Yoo, Kevin Yu, Qiming Yuan, Wojciech Zaremba, Rowan Zellers, Chong Zhang, Marvin Zhang, Shengjia Zhao, Tianhao Zheng, Juntang Zhuang, William Zhuk, and Barret Zoph. 2024. Gpt-4 technical report. *Preprint*, arXiv:2303.08774.

Kishore Papineni, Salim Roukos, Todd Ward, and Wei-Jing Zhu. 2002. Bleu: a method for automatic evaluation of machine translation. In *Proceedings of the 40th annual meeting on association for computational linguistics*, pages 311–318. Association for Computational Linguistics.

Vikas Raunak, Arul Menezes, Matt Post, and Hany Hassan. 2023. Do GPTs produce less literal translations? In *Proceedings of the 61st Annual Meeting of the Association for Computational Linguistics (Volume 2: Short Papers)*, pages 1041–1050, Toronto, Canada. Association for Computational Linguistics.

Ricardo Rei, Craig Stewart, Ana C Farinha, and Alon Lavie. 2020. Comet: A neural framework for mt evaluation. *Preprint*, arXiv:2009.09025.

Giancarlo Salton, Robert Ross, and John Kelleher. 2014. Evaluation of a substitution method for idiom transformation in statistical machine translation. In *Proceedings of the 10th Workshop on Multiword Expressions (MWE)*, pages 38–42, Gothenburg, Sweden. Association for Computational Linguistics.

Prateek Saxena and Soma Paul. 2020. Epie dataset: A corpus for possible idiomatic expressions. *Preprint*, arXiv:2006.09479.

David Stap, Eva Hasler, Bill Byrne, Christof Monz, and Ke Tran. 2024. The fine-tuning paradox: Boosting translation quality without sacrificing llm abilities. *Preprint*, arXiv:2405.20089.

Kenan Tang, Peiyang Song, Yao Qin, and Xifeng Yan. 2024. Creative and context-aware translation of east asian idioms with gpt-4. *Preprint*, arXiv:2410.00988.

NLLB Team, Marta R. Costa-jussà, James Cross, Onur Çelebi, Maha Elbayad, Kenneth Heafield, Kevin Heffernan, Elahe Kalbassi, Janice Lam, Daniel Licht,

Jean Maillard, Anna Sun, Skyler Wang, Guillaume Wenzek, Al Youngblood, Bapi Akula, Loic Barrault, Gabriel Mejia Gonzalez, Prangthip Hansanti, John Hoffman, Semarley Jarrett, Kaushik Ram Sadagopan, Dirk Rowe, Shannon Spruit, Chau Tran, Pierre Andrews, Necip Fazil Ayan, Shruti Bhosale, Sergey Edunov, Angela Fan, Cynthia Gao, Vedanuj Goswami, Francisco Guzmán, Philipp Koehn, Alexandre Mourachko, Christophe Ropers, Safiyyah Saleem, Holger Schwenk, and Jeff Wang. 2022. No language left behind: Scaling human-centered machine translation. *Preprint*, arXiv:2207.04672.

Qwen Team. 2024. Qwen2.5: A party of foundation models.

Zhangchen Xu, Fengqing Jiang, Luyao Niu, Yuntian Deng, Radha Poovendran, Yejin Choi, and Bill Yuchen Lin. 2024. Magpie: Alignment data synthesis from scratch by prompting aligned llms with nothing. *Preprint*, arXiv:2406.08464.

Masaru Yamada. 2024. Optimizing machine translation through prompt engineering: An investigation into chatgpt's customizability. *Preprint*, arXiv:2308.01391.

Tianyi Zhang, Varsha Kishore, Felix Wu, Kilian Q. Weinberger, and Yoav Artzi. 2020. Bertscore: Evaluating text generation with bert. In *International Conference on Learning Representations*.

Lianmin Zheng, Wei-Lin Chiang, Ying Sheng, Siyuan Zhuang, Zhanghao Wu, Yonghao Zhuang, Zi Lin, Zhuohan Li, Dacheng Li, Eric P. Xing, Hao Zhang, Joseph E. Gonzalez, and Ion Stoica. 2023. Judging llm-as-a-judge with mt-bench and chatbot arena. In *Advances in Neural Information Processing Systems 36: Annual Conference on Neural Information Processing Systems 2023, NeurIPS 2023, New Orleans, LA, USA, December 10 - 16, 2023*.

Yafei Zhu, Daisy Monika Lal, Sofiia Denysiuk, and Ruslan Mitkov. 2024. From neural machine translation to large language models: Analysing translation quality of chinese idioms. *Proceedings of the International Conference on New Trends in Translation and Technology Conference 2024*.